\documentclass[runningheads]{llncs}
\usepackage{graphicx}
\usepackage{epsfig}
\usepackage{url}
\usepackage{bbm}
\usepackage{commath}
\usepackage{algorithm,algpseudocode}
\usepackage{booktabs}
\usepackage{amsmath,amssymb}
\usepackage[export]{adjustbox}
\usepackage{color}
\DeclareMathOperator*{\argmin}{arg\,min}
\DeclareMathOperator*{\argmax}{arg\,max}

\begin{document}
\title{Towards Robust Neural Networks via Random Self-ensemble} 

\titlerunning{RSE for Robust Neural Networks}
\author{Xuanqing Liu\inst{1} \and
Minhao Cheng\inst{1} \and
Huan Zhang\inst{2} \and
Cho-Jui Hsieh\inst{1,3}
}
\authorrunning{X. Liu, M. Cheng, H. Zhang and C-J. Hsieh}
% Replace with shorter version of the author list. If there are more authors than fits a line, please use A. Author et al.
%

\institute{Electrical and Computer Science, UC Davis, Davis CA 95616, USA
\\\email{\{xqliu, mhcheng\}@ucdavis.edu}
\and
Electrical and Computer Engineering, UC Davis, Davis CA 95616, USA
\\\email{ecezhang@ucdavis.edu}
\and
Department of Statistics, UC Davis, Davis CA 95616, USA\\
\email{chohsieh@ucdavis.edu}}
\maketitle              % typeset the header of the contribution
\begin{abstract}
Recent studies have revealed the vulnerability of deep neural networks: A small adversarial perturbation that is imperceptible to human can easily make a well-trained deep neural network misclassify. This makes it unsafe to apply neural networks in security-critical applications. In this paper, we propose a new defense algorithm called Random Self-Ensemble (RSE) by combining two important concepts: {\bf randomness} and {\bf ensemble}. To protect a targeted model, RSE adds random noise layers to the neural network to prevent the strong gradient-based attacks, and ensembles the prediction over random noises to stabilize the performance. We show that our algorithm is equivalent to ensemble an infinite number of noisy models $f_\epsilon$ without any additional memory overhead, and the proposed training procedure based on noisy stochastic gradient descent can ensure the ensemble model has a good predictive capability. Our algorithm significantly outperforms previous defense techniques on real data sets. For instance, on CIFAR-10 with VGG network (which has 92\% accuracy without any attack), under the strong C\&W attack within a certain distortion tolerance, the accuracy of unprotected model drops to less than 10\%, the best previous defense technique has $48\%$ accuracy, while our method still has $86\%$ prediction accuracy under the same level of attack. Finally, our method is simple and easy to integrate into any neural network.

\end{abstract}

\section{Introduction}
Deep neural networks have demonstrated their success in many machine learning and computer vision applications, including image classification~\cite{he2016deep,dean2012large,xiao2018spatially,eykholt2018robust,ijcai2018-543}, object recognition~\cite{szegedy2015going} and image captioning~\cite{xu2015show}. Despite having near-perfect prediction performance, recent studies have revealed the vulnerability of deep neural networks to adversarial examples---given a correctly classified image, a carefully designed perturbation to the image can make a well-trained neural network misclassify. Algorithms crafting these adversarial images, called attack algorithms, are designed to minimize the perturbation, thus making adversarial images hard to be distinguished from natural images. This leads to security concerns, especially when applying deep neural networks to security-sensitive systems such as self-driving cars and medical imaging. 
\par
To make deep neural networks more robust to adversarial attacks, several defense algorithms have been proposed recently~\cite{papernot2016practical,zantedeschi2017efficient,kurakin2016adversarial,huang2015learning,xu2017feature}. However, recent studies showed that these defense algorithms can only marginally improve the accuracy under the adversarial attacks~\cite{carlini2017adversarial,carlini2017towards}. 
\par
In this paper, we propose a new defense algorithm: Random Self-Ensemble (RSE). 
More specifically, we introduce the new ``noise layer'' that fuses input vector with randomly generated noise, and then we insert this layer before each convolution layer of a deep network. In the training phase, the gradient is still computed by back-propagation but it will be perturbed by random noise when passing through the noise layer. In the inference phase, we perform several forward propagations, each time with different prediction scores due to the noise layers, and then ensemble the results. We show that RSE makes the network more resistant to adversarial attacks, by virtue of the proposed training and testing scheme. Meanwhile, it will only slightly affect test accuracy when no attack is performed on natural images. The algorithm is trivial to implement and can be applied to any deep neural networks for the enhancement. 
\par
Intuitively, RSE works well because of two important concepts: {\bf ensemble} and {\bf randomness}. It is known that ensemble of several trained models can improve the robustness~\cite{ensemble}, but will also increase the model size by $k$ folds.  
In contrast, without any additional memory overhead, RSE can construct infinite number of models $f_\epsilon$, where $\epsilon$ is generated randomly, and then ensemble the results to improve robustness. But how to guarantee that the ensemble of these models can achieve good accuracy? After all, if we train the original model without noise, yet only add noise layers at the inference stage, the algorithm is going to perform poorly. This suggests that adding random noise to an pre-trained network will only degrade the performance. Instead, we show that if the noise layers are taken into account in the training phase, then the training procedure can be considered as minimizing the upper bound
of the loss of model ensemble, and thus our algorithm can achieve good prediction accuracy.
\par
The contributions of our paper can be summarized below: 
\begin{itemize}
    \item We propose the Random Self-Ensemble (RSE) approach for improving the robustness of deep neural networks. The main idea is to add a ``noise layer'' before each convolution layer in both training and prediction phases. The algorithm is equivalent to ensemble an infinite number of random models to defense against the attackers. 
    \item We explain why RSE can significantly improve the robustness toward 
    adversarial attacks and show that adding noise layers is equivalent to training the original network with an extra regularization of Lipschitz constant. 
    \item RSE significantly outperforms existing defense algorithms in all our experiments. For example, on CIFAR-10 data and VGG network (which has 92\% accuracy without any attack), under C\&W attack the accuracy of unprotected model drops to less than 10\%; the best previous defense technique has $48\%$ accuracy; while RSE still has $86.1\%$ prediction accuracy under the same strength of attacks. Moreover, RSE is easy to implement and can be combined with any neural network.
\end{itemize}

\section{\label{sec:related}Related Work}
Security of deep neural networks has been studied recently. Let us denote the neural network as $f(w, x)$ where $w$ is the model parameters and $x$ is the input image. Given a correctly classified image $x_0$ ($f(w, x_0)=y_0$), an attacking algorithm seeks to find a slightly perturbed image $x'$ such that: (1) the neural network will misclassify this perturbed image; and (2) the distortion $\|x'-x_0\|$ is small so that the perturbation
is hard to be noticed by human. A defense algorithm is designed to improve the robustness of neural networks against attackers, usually by slightly changing the loss function or training procedure. In the following, we summarize some recent works along this line. 

\subsection{White-box attack}
In the white-box setting, attackers have all information about the targeted neural network, including network structure and network weights (denoted by $w$). 
Using this information, attackers can compute gradient with respect to input data $\nabla_x f(w, x)$ by back-propagation. Note that gradient is very informative for attackers since it characterizes the sensitivity of the prediction with respect to the input image. 
\par
To craft an adversarial example, \cite{goodfellow2014explaining} proposed
a fast gradient sign method (FGSM), where the adversarial example is constructed by 
\begin{equation}
    x' = x_0 - \epsilon\cdot \text{sign}(\nabla_x f(w, x_0))
\end{equation}
with a small $\epsilon>0$. Based on that, several followup works were made to improve the efficiency and availability, such as Rand-FGSM~\cite{tramer2017ensemble} and I-FGSM~\cite{kurakin2016adversarial}. Recently, Carlini \& Wagner~\cite{carlini2017towards} showed that constructing an adversarial example can be formulated as solving the following optimization problem: 
\begin{equation}
x' = \min_{x\in[0,1]^d} c\cdot g(x) + \|x-x_0\|^2_2, 
\label{eq:c_w}
\end{equation}
where the first term is the loss function that characterizes the success of the attack and
the second term is to enforce a small distortion. The parameter $c>0$ is used
to balance these two requirements. Several variants were proposed recently~\cite{chen2017ead,madry2017towards}, but most of them can be categorized in the similar framework. The C\&W attack has been recognized as a strong attacking
algorithm to test defense methods.
\par
For untargeted attack, where the goal is to find an adversarial example that is close to the original example but yields different class prediction, the loss function in~\eqref{eq:c_w} can be defined as
\begin{equation}
g(x) = \max\{\max_{i\neq t}(Z(x')_i)-Z(x')_t, -\kappa\},
\end{equation}
where $t$ is the correct label, $Z(x)$ is the network's output before softmax layer (logits). 
\par
For targeted attack, the loss function can be designed to force the classifier to return the target label. For attackers, targeted attack is strictly harder than untargeted attack (since once the targeted attack succeeds, the same adversarial image can be used to perform untargeted attack without any modification). On the contrary, for defenders, untargeted attacks are strictly harder to defense than targeted attack. Therefore, 
we focus on defending the untargeted attacks in our experiments. 
%Note that even if the prediction model is an ensemble, we can still find the adversarial images efficiently, the details are delayed into appendix.

\subsection{Defense Algorithms}
Because of the vulnerability of adversarial examples~\cite{szegedy2013intriguing}, several methods have been proposed to improve the network's robustness against adversarial examples. \cite{papernot2016distillation} proposed \emph{defensive distillation}, which uses a modified softmax layer controlled by temperature to train the ``teacher'' network, and then use the prediction probability (soft-labels) of teacher network to train the student network (it has the same structure as the teacher network). However, as stated in \cite{carlini2017towards}, this method does not work properly when dealing with the C\&W attack. Moreover, \cite{zantedeschi2017efficient} showed that by using a modified ReLU activation layer (called BReLU) and adding noise into origin images to augment the training dataset, the learned model will gain some stability to adversarial images. 
Another popular defense approach is \emph{adversarial training}~\cite{kurakin2016adversarial,huang2015learning}. It generates and appends adversarial examples found by an attack algorithm to the training set, which helps the network to learn how to distinguish adversarial examples. Through combining adversarial training with enlarged model capacity, \cite{madry2017towards} is able to create an MNIST model that is robust to the first order attacks, but this approach does not work very well on more difficult datasets such as CIFAR-10. 
\par
It is worth mentioning that there are many defense algorithms (\textit{r.f.} \cite{buckman2018thermometer,ma2018characterizing,guo2018countering,s.2018stochastic,xie2018mitigating,song2018pixeldefend,samangouei2018defensegan}) against white box attacks in literature. Unfortunately, as \cite{athalye2018obfuscated,athalye2018robustness} pointed out, these algorithms are not truly effective to white box attacks. Recall the ``white box'' means that the attackers know \textit{everything} concerning how models make decisions, these include the potential defense mechanisms. In this condition, the white box attacks can walk around all defense algorithms listed above and the accuracy under attack can still be nearly zero. In addition to changing the network structure, there are other methods ~\cite{xu2017feature,meng2017magnet,feinman2017detecting,grosse2017statistical} ``detecting'' the adversarial examples, which are beyond the scope of our paper. 
\par
There is another highly correlated work (\textit{r.f.} \cite{2018arXiv180203471L}) which also adopts very similar idea, except that they view this problem from the angle of differential privacy, while we believe that the adversarial robustness is more correlated with regularization and ensemble learning. Furthermore, our work is public available earlier than this similar work on Arxiv. 
\section{Proposed Algorithm: Random Self-Ensemble}
\label{sec:proposed}
\begin{figure*}[tb]
    \centering
    \includegraphics[width=0.95\linewidth]{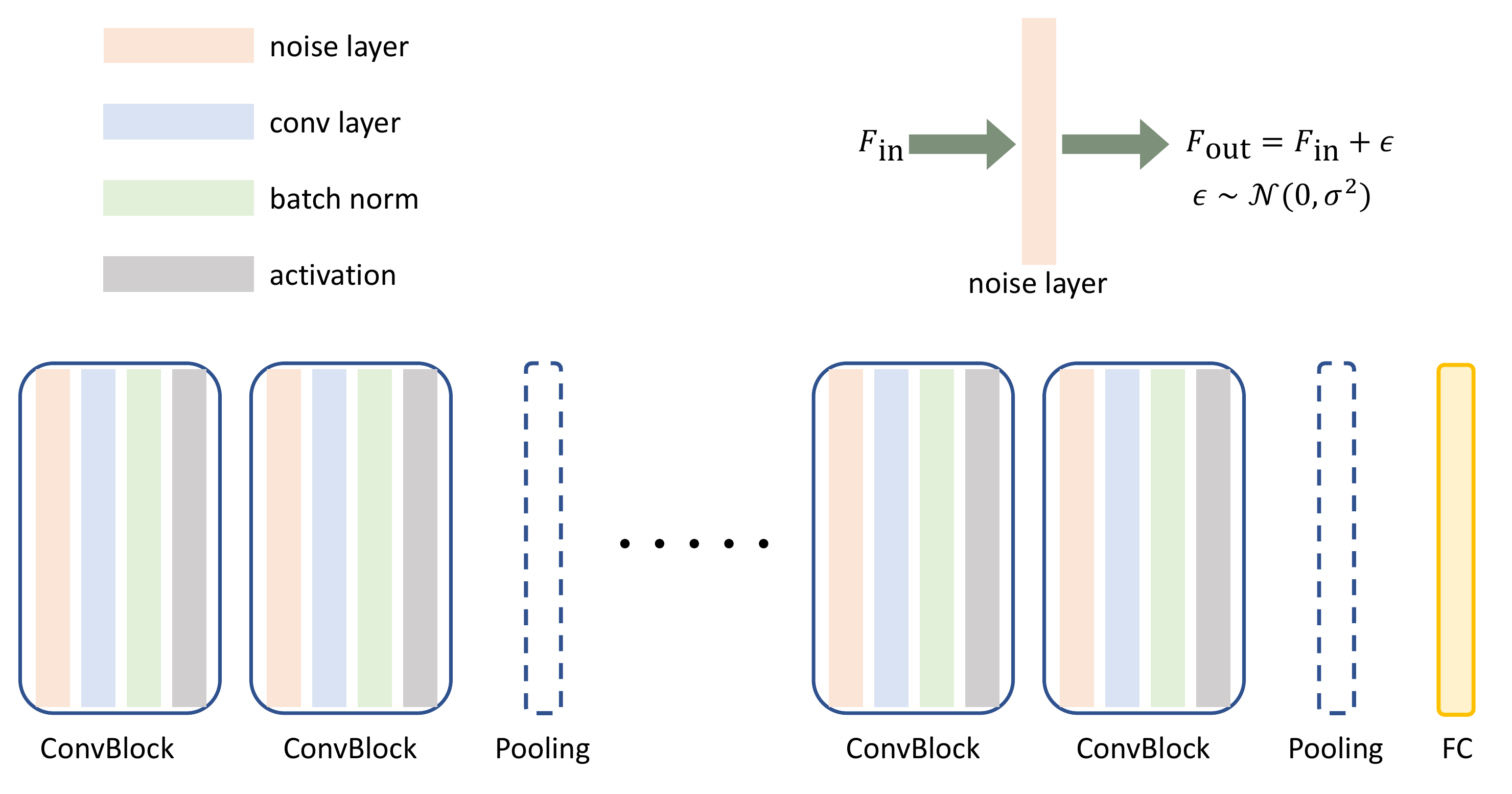}
    \caption{Our proposed noisy VGG style network, we add a noise layer before each convolution layer. For simplicity, we call the noise layer before the first convolution layer the ``init-noise'', and all other noise layer ``inner-noise''. For these two kinds of layers we adopt different variances of Gaussian noise. Note that similar design can be transplanted to other architectures such as ResNet.}
    \label{fig:vgg}
\end{figure*}

In this section, we propose our self-ensemble algorithm to improve the robustness of neural networks. 
We will first motivate and introduce our algorithm and then discuss several theoretical reasons behind it.
%our algorithm.
%and why it can improve the robustness under
%various white-box or black-box attacks.
%Our algorithm is very simple and can be easily applied to any existing networks. 

It is known that ensemble of several different models can improve the robustness.
However, an ensemble of finite $k$ models is not very practical because it will increase the model size by
$k$ folds. 
For example, AlexNet model on ImageNet requires 240MB storage, and storing 100 of them will require 24GB memory.
Moreover, it is hard to find many heterogeneous models with similar accuracy. 
To improve the robustness of practical systems, we propose the following self-ensemble algorithm that
can generate an infinite number of models on-the-fly without any additional memory cost. 
%and furthremore the models are random so it is hard for attackers to attack any of them. 
%Can we generate infinite number of models on-the-fly, without any additional memory cost, and the models are random so it's hard for attackers to attack any of them. 

Our main idea is to add randomness into the network structure. More specifically,
we introduce a new ``noise layer'' that fuses input vector with a randomly generated noise, 
 i.e. $x\rightarrow x+\epsilon$ when passing through the noise layer. Then we 
add this layer before each convolution layer as shown in Fig.~\ref{fig:vgg}. Since most attacks
require computing or estimating gradient, the noise level in our model will control the success rate of those attacking algorithms. 
In fact, we can integrate this layer into any other neural network.
%to acquire randomness. 

If we denote the original neural network as $f(w, x)$ where $w\in\mathbb{R}^{d_w}$ is the weights and $x\in\mathbb{R}^{d_x}$ is the input image, then considering the random noise layer, the network can be denoted as $f_{\epsilon}(w, x)$ with random $\epsilon\in\mathbb{R}^{d_e}$. 
%With this layer, the network becomes stochastic - each time the network can be denoted as $f_\epsilon$
%with a different random number $\epsilon$, 
Therefore we have an infinite number of models in the pocket (with different $\epsilon$) without having 
any memory overhead. However, adding randomness will also affect the prediction accuracy of the model. How can we make sure that the ensemble of these random models have enough accuracy? 
%we add randomness in each layer of neural network XXX (add xuanqing's paragraph here). With randomness, each time the model will be different, so we have infinite models in the pocket but there is no additional cost.  However, how can we guarantee the ensemble of these random models are good? 
%we introduce a new layer that helps stabilize neural networks and greatly improves the accuracy under adversarial attacks, the basic idea is to add noise to input data before each convolution layer as shown in Figure~\ref{fig:vgg}, 

\begin{algorithm}[tb]
\caption{\label{alg:train}Training and Testing of Random Self-Ensemble (RSE)}
\begin{algorithmic}
\State \textbf{Training phase}: 
\For{ $\text{iter}=1, 2, \dots$ %$(x_i, y_i)$ in dataset
}
\State Randomly sample $(x_i, y_i)$ in dataset
\State Randomly generate $\epsilon\!\sim\!\mathcal{N}(0, \sigma^2)$ for each noise layer.
\State Compute $\Delta w=\nabla_w \ell(f_\epsilon(w, x_i), y_i)$
(Noisy gradient)
%\State Forward propagation: $\ell\big(f_{\epsilon}(w, x_i), y_i\big)$.
\State Update weights: $w\leftarrow w - \Delta w$.
\EndFor
\State \textbf{Testing phase}:
\State Given testing image $x$, initialize $p=(0, 0,\dots, 0)$
\For{$j=1,2,\dots, \text{\#Ensemble}$}
\State Randomly generate $\epsilon\!\sim\!\mathcal{N}(0, \sigma^2)$ for each noise layer.
\State Forward propagation to calculate probability output
$$p^j=f_\epsilon(w, x)$$
\State Update $p$: $p\leftarrow p+p^j$.
\EndFor
\State Predict the class with maximum score $\hat{y}=\argmax_k p_k$
\end{algorithmic}
\end{algorithm}

A critical observation is that we need to add this random layer in both training and testing phases. 
The training and testing algorithms are listed in Algorithm~\ref{alg:train}. 
In the training phase, gradient is computed as $\nabla_w f_\epsilon(w, x_i)$ which includes the noise layer, 
and the noise is generated randomly for each stochastic gradient descent update. In the testing phase, 
we construct $n$ random noises and ensemble their probability outputs by 
\begin{equation}
    p = \sum_{j=1}^n f_{\epsilon_j}(w, x), \text{ and predict } \hat{y} = \arg\max_k p_k. 
\end{equation}
If we do not care about the prediction time, $n$ can be very large, but in practice we found it saturates at $n\approx 10$ (see Fig.~\ref{fig:n-ensemble}). 

%is e. 
This approach is different from Gaussian data augmentation in ~\cite{zantedeschi2017efficient}: they only add Gaussian noise to images during the training time, while we add noise before each convolution layer at both training and inference time. When training, the noise helps optimization algorithm to find a stable convolution filter that is robust to perturbed input, while when testing, the roles of noise are two-folded: one is to perturb the gradient to fool gradient-based attacks.%so much that gradient based attacks are affected.
The other is it gives different outputs by doing multiple forward operations and a simple ensemble method can improve the testing accuracy. 
%This ensembling method differs from commonly seen multi-model ensembling 
%in the aspect that we only need to store one model and forward multiple times, thus we call our algorithm ``self-ensembling''. 

\subsection{Mathematical explanations}

\paragraph{Training and testing of RSE. }
Here we explain our training and testing procedure. 
%If we denote the network as $f(w, x)$ where $w$ is the weights and $x$ is the input image, and $f_{\epsilon}(w, x)$ is the perturbed network by Gaussian noise $\epsilon$, then 
In the training phase, our algorithm is solving the following 
optimization problem: 
%the training process of our method is
\begin{equation}
    \label{eq:train-noise-net}
    \begin{aligned}
    w^*&= \argmin_{w}\frac{1}{|\mathcal{D}_{\text{train}}|}\sum_{(x_i, y_i)\in\mathcal{D}_{\text{train}}}\mathop{\mathbb{E}}_{\epsilon\sim \mathcal{N}(0, \sigma^2)}\ell\big(f_{\epsilon}(w,x_i), y_i\big),
    \end{aligned}
\end{equation}
where $\ell(\cdot, \cdot)$ is the loss function and  $\mathcal{D}_{\text{train}}$ is the training dataset.
Note that for simplicity we assume $\epsilon$ follows a zero-mean Gaussian, but in general our algorithm
can work for a large variety of noise distribution such as Bernoulli-Gaussian: $\epsilon_i=b_ie_i$, where $e_i\overset{\text{iid}}{\sim}\mathcal{N}(0, \sigma^2)$ and $b_i\overset{\text{iid}}{\sim}\mathcal{B}(1, p)$.
\par
At testing time, we ensemble the outputs through several forward propagation, specifically:
\begin{equation}
    \label{eq:ensemble-infer}
        \hat{y}_i=\argmax\mathbb{E}_{\epsilon\sim \mathcal{N}(0, \sigma^2)} f_{\epsilon}(w, x_i). 
\end{equation}
Here~$\argmax$~means the index of maximum element in a vector. The reason that our RSE algorithm achieves the similar prediction accuracy with original network
is because 
%we should note that 
\eqref{eq:train-noise-net} is minimizing an upper bound of the loss of \eqref{eq:ensemble-infer} -- Similar to the idea of \cite{noh2017regularizing}, if we choose negative log-likelihood loss, then $\forall w\in\mathbb{R}^{d_w}$:% and $(x_i, y_i)\sim\mathcal{P}_{\text{data}}$: 
\begin{equation}
    \label{eq:bound-loss}
    \begin{aligned}
    &\frac{1}{|\mathcal{D}_{\text{train}}|}\sum_{(x_i,y_i)\in\mathcal{D}_{\text{train}}}\mathbb{E}_{\epsilon\sim\mathcal{N}(0, \sigma^2)}\ell\big(f_{\epsilon}(w, x_i), y_i\big)\\
    &\quad\overset{(a)}{\approx}\mathbb{E}_{(x_i, y_i)\sim\mathcal{P}_{\text{data}}}\Big\{-\mathbb{E}_{\epsilon\sim\mathcal{N}(0, \sigma^2)}\log f_{\epsilon}(w, x_i)[y_i]\Big\}\\
    &\quad\overset{(b)}{\ge} \mathbb{E}_{(x_i, y_i)\sim\mathcal{P}_{\text{data}}}\Big\{-\log\mathbb{E}_{\epsilon\sim\mathcal{N}(0, \sigma^2)}f_{\epsilon}(w, x_i)[y_i]\Big\}\\
    &\quad\overset{(c)}{\ge} \mathbb{E}_{(x_i, y_i)\sim\mathcal{P}_{\text{data}}}\Big\{-\log\mathbb{E}_{\epsilon\sim\mathcal{N}(0, \sigma^2)}f_{\epsilon}(w, x_i)[\hat{y}_i]\Big\}\\
    &\quad\overset{(a)}{\approx}\frac{1}{|\mathcal{D}_{\text{test}}|}\sum_{x_i\in\mathcal{D}_{\text{test}}}-\log\mathbb{E}_{\epsilon\sim\mathcal{N}(0, \sigma^2)}f_{\epsilon}(w, x_i)[\hat{y}_i].
    \end{aligned}
\end{equation}
Where $\mathcal{P}_{\text{data}}$ is the data distribution, $\mathcal{D}_{\text{train/test}}$ is the training set and test set, respectively. And $(a)$ follows from generalization bound (see~\cite{steinhardt2017certified} or appendix for details), $(b)$ comes from Jensen's inequality and $(c)$ is by the inference rule \eqref{eq:ensemble-infer}. So by minimizing \eqref{eq:train-noise-net} we are actually minimizing the upper bound of inference confidence $-\log f_{\epsilon}(w, x_i)[\hat{y}_i]$, this validates our ensemble inference procedure.
\par
\paragraph{RSE is equivalent to Lipschitz regularization. }

Another point of view is that perturbed training is equivalent to Lipschitz regularization, which further helps defensing gradient based attack. If we fix the output label $y$ then the loss function $\ell(f_\epsilon(w, x), y)$ can be simply denoted as $\ell \circ f_{\epsilon}$. Lipchitz of the function $\ell\circ f_{\epsilon}$ is a constant $L_{\ell\circ f_{\epsilon}}$ such that
%
%For input image $x$, we have attacking image $\tilde{x}$ such that
\begin{equation}
    \label{eq:lipschitz}
    |\ell(f_{\epsilon}(w,x),y)-\ell(f_{\epsilon}(w,\tilde{x}),y)|\le L_{\ell\circ f_{\epsilon}}\|x-\tilde{x}\|
\end{equation}
for all $x, \tilde{x}$. 
%Imaging $x$ is the original image and $\tilde{x}$ is the adversarial image, 
%then $L_{\ell\circ f}$ characterizes the sensitivity of loss with respect to input perturbation. 
%where $L_{\ell\circ f}$ is the Lipschitz of composite function $\ell\circ f$. 
In fact, it has been proved recently that Lipschitz constant can be used to measure the robustness
of machine learning model~\cite{hein2017formal,weng2018evaluating}. 
If $L_{\ell\circ f_{\epsilon}}$ is large enough, even a tiny change of input $x-\tilde{x}$ can significantly change the loss and eventually get an incorrect prediction. On the contrary,
by controlling $L_{\ell\circ f}$ to be small, we will have a more robust network. 

Next we show that our noisy network indeed controls the Lipschitz constant. Following the notation of \eqref{eq:train-noise-net}, we can see that 
\begin{equation}
    \label{eq:lipschitz-regularize}
    \begin{aligned}
    \mathbb{E}_{\epsilon\sim\mathcal{N}(0,\sigma^2)}\ell\big(f_{\epsilon}(w,x_i), y_i\big)&\overset{(a)}{\approx}\mathbb{E}_{\epsilon\sim\mathcal{N}(0,\sigma^2)}\Big[\ell\big(f_0(w, x_i), y_i\big)+\epsilon^{\intercal}\nabla_\epsilon \ell\big(f_{0}(w, x_i), y_i\big)\\
    &\qquad+\frac{1}{2}\epsilon^\intercal \nabla_\epsilon^2 \ell\big(f_0(w, x_i), y_i\big)\epsilon\Big]\\
    &\overset{(b)}{=} \ell\big(f_0(w, x_i), y_i\big)+\frac{\sigma^2}{2}\text{Tr}\Big\{\nabla_\epsilon^2\ell\big(f_0(w, x_i), y_i\big)\Big\}.
    \end{aligned}
\end{equation}
For $(a)$, we do Taylor expansion at $\epsilon=0$. Since we set the variance of noise $\sigma^2$ very small, we only keep the second order term. For $(b)$, we notice that the Gaussian vector $\epsilon$ is i.i.d. with zero mean. So the linear term of $\epsilon$ has zero expectation, and the quadratic term is directly dependent on variance of noise and the trace of Hessian. As a convex relaxation, if we assume $\ell\circ f_0$ is convex, then we have that $d\cdot \|A\|_{\max}\ge \text{Tr}(A) \ge \|A\|_{\max}$ for $A\in\mathbb{S}_{+}^{d\times d}$, we can rewrite \eqref{eq:lipschitz-regularize} as
\begin{equation}
    \label{eq:loss-regularized}
    \text{Loss}(f_\epsilon, \{x_i\}, \{y_i\})\simeq\text{Loss}(f_0, \{x_i\}, \{y_i\})+\frac{\sigma^2}{2}L_{\ell\circ f_0},
\end{equation}
which means the training of noisy networks is equivalent to training the original model with an extra regularization of Lipschitz constant, and by controlling the variance of noise we can balance the robustness of network with training loss.

\subsection{Discussions}

Here we show both {\it randomness} and {\it ensemble} are important in our algorithm. 
Indeed, if we remove any component, the performance will significantly drop. 
%And some naive ways to add random noise and ensemble does not work. 

\begin{figure}[htb]
    \centering
    \includegraphics[width=0.5\linewidth]{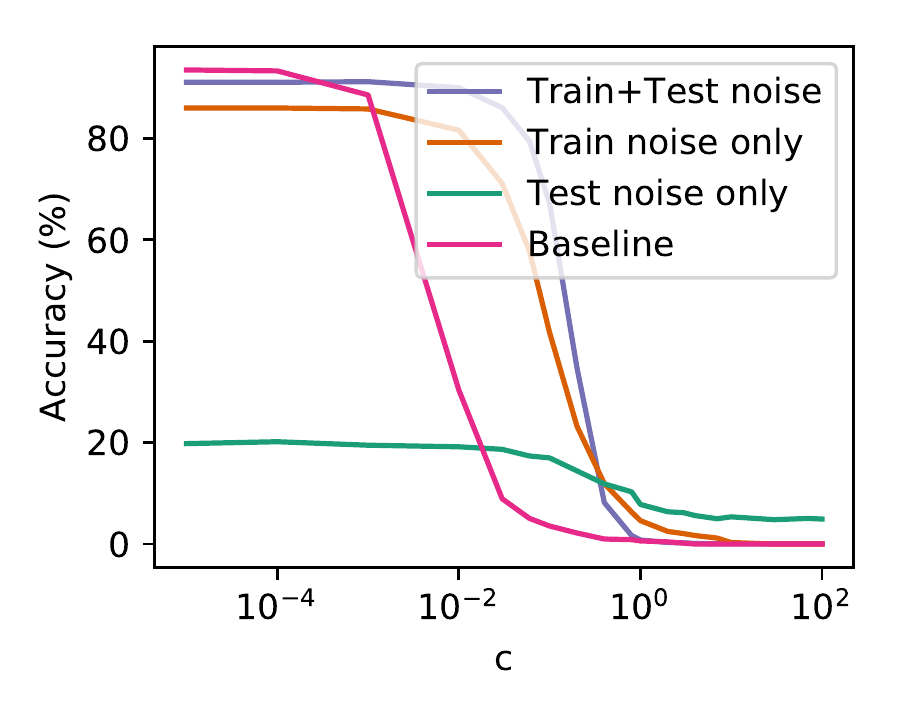}
    \caption{We test three models on CIFAR10 and VGG16 network: In the first model noise is added both at training and testing time, in the second model noise is added only at training time, in the last model we only add noise at testing time. As a comparison we also plot baseline model which is trained conventionally. For all models that are noisy at testing time, we automatically enable self-ensemble.}
    \label{fig:noise_ensemble}
\end{figure}

First, as mentioned before, the main idea of our model is to have infinite number of models $f_\epsilon$, each
with a different $\epsilon$ value, and then ensemble the result. A naive way to achieve
this goal is to fix a pre-trained model $f_0$ and then generate many $f_\epsilon$ in the testing
phase by adding different small noise to $f_0$. However, Fig.~\ref{fig:noise_ensemble} shows this approach (denoted as Test noise only) will result in
much worse performance (20\% without any attack). Therefore it is non-trivial to guarantee the model
to be good after adding small random noise. In our random self-ensemble algorithm, in addition to adding noise in the testing phase, 
we also {\bf add noise layer in the training phase}, and this is important for getting good performance. 
\par
%Second, there are several existing approaches try to add noise in the training phase but not testing phase \cite{zantedeschi2017efficient}. 
Second, we found adding noise in the testing phase and then ensemble the predictions is important. In Fig.~\ref{fig:noise_ensemble}, we compare the performance of RSE with 
the version that only adds the noise layer in the training phase but not in the testing phase (so the prediction is $f_{\epsilon}(w, x)$
instead of $\mathbb{E}_\epsilon f_\epsilon(w, x)$). The results clearly show that the performance drop under smaller attacks. This proves {\bf ensemble in the testing phase is crucial}.
%Note that if we only add the noise layer in the input layer
%and do not do ensemble in the testing phase, the algorithm is equivalent to the robust optimization approach \cite{zantedeschi2017efficient}. In this sense, our method can be regarded as a natural extension of robust optimization. 

%\subsection{Resistant against black-box attack (ZOO)}
%ZOO~\cite{chen2017zoo} is the most successful attack algorithm in the black-box setting and outperforms transfer attacks by a significant amount (see~\cite{chen2017zoo}). Interestingly, the accuracy of ZOO attack is theoretically controlled by the noise $\epsilon$ added in our noise layer. Recall that ZOO crafts the adversarial example by solving the optimization problem similar to~\eqref{eq:c_w} using zero-th order optimization, where gradient is estimated by finite difference. However, with RSE, the gradient estimator computed by ZOO will become
%\begin{equation}
%    \frac{f_{\epsilon_1}(x+\delta e_i) - f_{\epsilon_2}(x - %\delta e_i)}{2\delta}, 
%\end{equation}
%which is no longer an estimator of $\nabla f(x)$ even when $\delta\rightarrow 0$ because $f_{\epsilon_1} \neq f_{\epsilon_2}$. Therefore, ZOO will not even converge.
\section{Experiments}
\paragraph{Datasets and network structure}
We test our method on two datasets---CIFAR10 and STL10. We do not compare the results on MNIST since it is a much easier dataset and existing defense methods such as ~\cite{papernot2016practical,zantedeschi2017efficient,kurakin2016adversarial,huang2015learning} can effectively increase image distortion under adversarial attacks. On CIFAR10 data, we evaluate the performance on both VGG-16~\cite{simonyan2014very} and ResNeXt \cite{xie2016aggregated}; on STL10 data we copy and slightly modify a simple model\footnote{Publicly available at \url{https://github.com/aaron-xichen/pytorch-playground}} which we name it as ``Model A''.

\paragraph{Defense algorithms.} We include the following defense algorithms into comparison (their parameter settings can be found in Tab.~\ref{tab:compare-set}): 
\begin{itemize}
    \item Random Self-Ensemble (RSE): our proposed method.
    \item Defensive distillation \cite{papernot2016distillation}: first train a teacher network at temperature $T$, then use the teacher network to train a student network of the same architecture and same temperature. The student network is called the distilled network.
    \item Robust optimization combined with BReLU activation~\cite{zantedeschi2017efficient}: first we replace all ReLU activation with BReLU activation. And then at the training phase, we randomly perturb training data by Gaussian noise with $\sigma=0.05$ as suggested.
    \item Adversarial retraining by FGSM attacks~\cite{kurakin2016adversarial,huang2015learning}: we first pre-train a neural network without adversarial retraining. After that, we either select an original data batch or an adversarial data batch  with probability $1/2$. We continue training it until convergence.
\end{itemize} 

\paragraph{Attack models.} 
%Although nowadays there are many attacking methods discussed in Section~\ref{sec:related}, they differ greatly on the power of attacks. Obviously white-box attacks know more information about the targeted model so they have higher success rate, qualifying itself as a challenger to defense models. 
We consider the white-box setting and choose the state-of-the-art C\&W attack~\cite{carlini2017towards} to evaluate the above-mentioned defense methods. 
Moreover, we test our algorithm under untargeted attack, since untargeted attack is strictly harder to defense than targeted attack. 
In fact, C\&W untargeted attack is the most challenging attack for a defense algorithm. 
%As experiment in ~\cite{carlini2017towards} shows, C\&W attack should be the benchmark for defense methods.

Moreover, we assume C\&W attack knows the randomization procedure of RSE, so the C\&W objective function will change accordingly (as proposed in~\cite{athalye2018robustness} for attacking an ensemble model). The details can be found in the appendix. 

\paragraph{Measure.} Unlike attacking models that only need to operate on correctly classified images, a competitive defense model not only protects the model when attackers exist, but also keeps a good performance on clean datasets. Based on this thought, we compare the accuracy of guarded models under different strengths of C\&W attack, the strength can be measured by $L_2$-norm of image distortion and further controlled by parameter $c$ in~\eqref{eq:c_w}.
Note that an adversarial image is correctly predicted under C\&W attack if and only if 
the original image is correctly classified and C\&W attack cannot find an adversarial example within a certain distortion level. 

\begin{figure}[h!]
    \centering
    \includegraphics[width=0.45\linewidth]{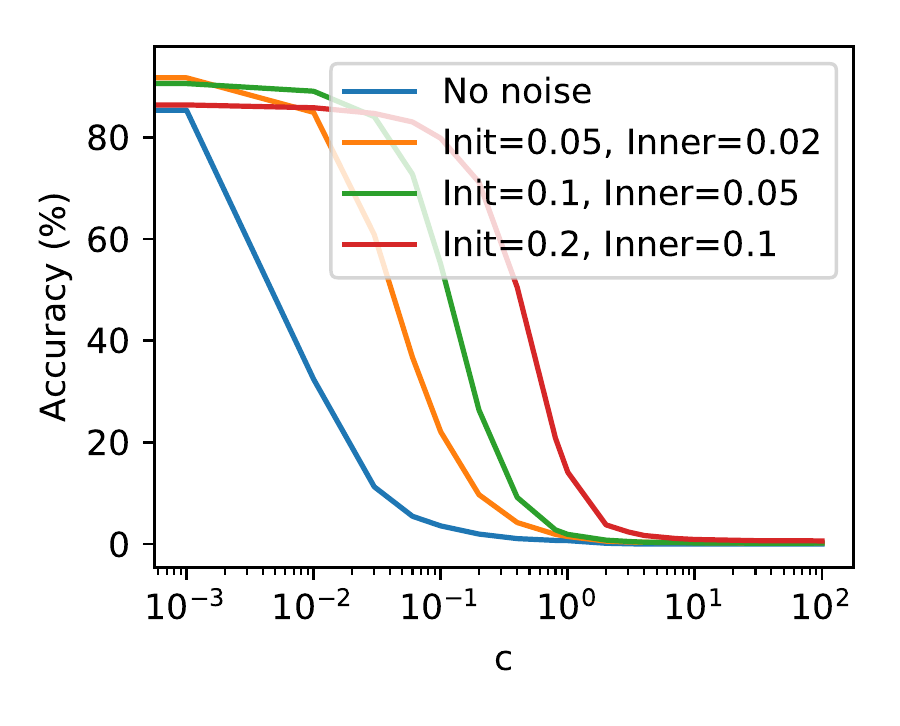}
    \includegraphics[width=0.45\linewidth, height=0.363\linewidth]{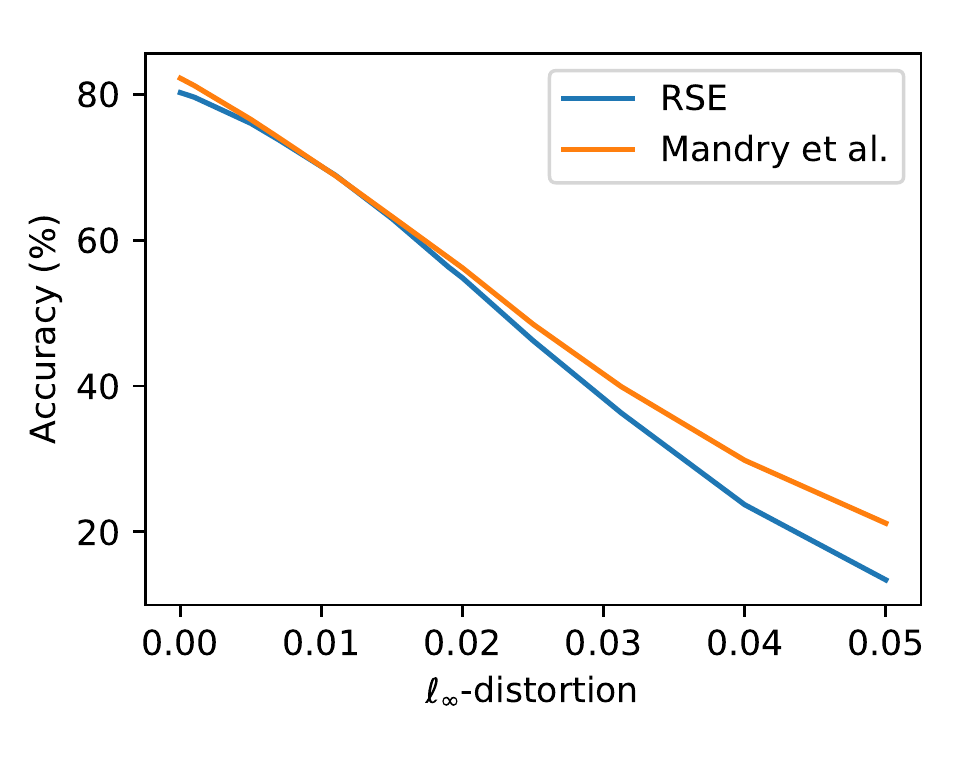}
    \caption{\label{fig:noise-level-robust}\textit{Left}: the effect of noise level on robustness and generalization ability. Clearly random noise can improve the robustness of the model. \textit{Right}: comparing RSE with adversarial defense method~\cite{madry2017towards}.}
\end{figure}
\begin{figure}[tb]
    \centering
    \includegraphics[width=0.45\linewidth]{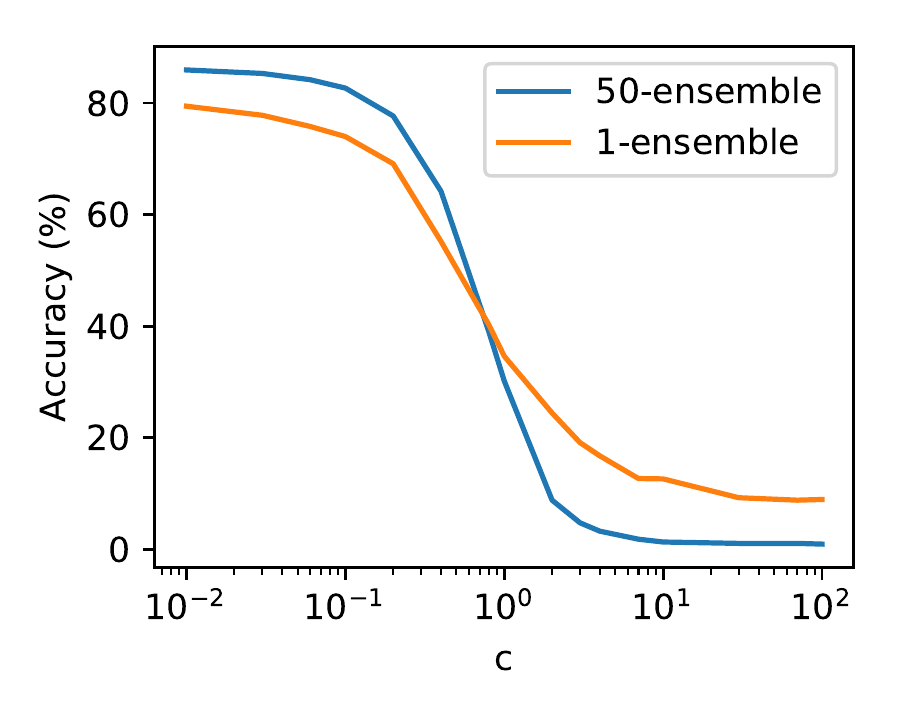}
    \includegraphics[width=0.45\linewidth]{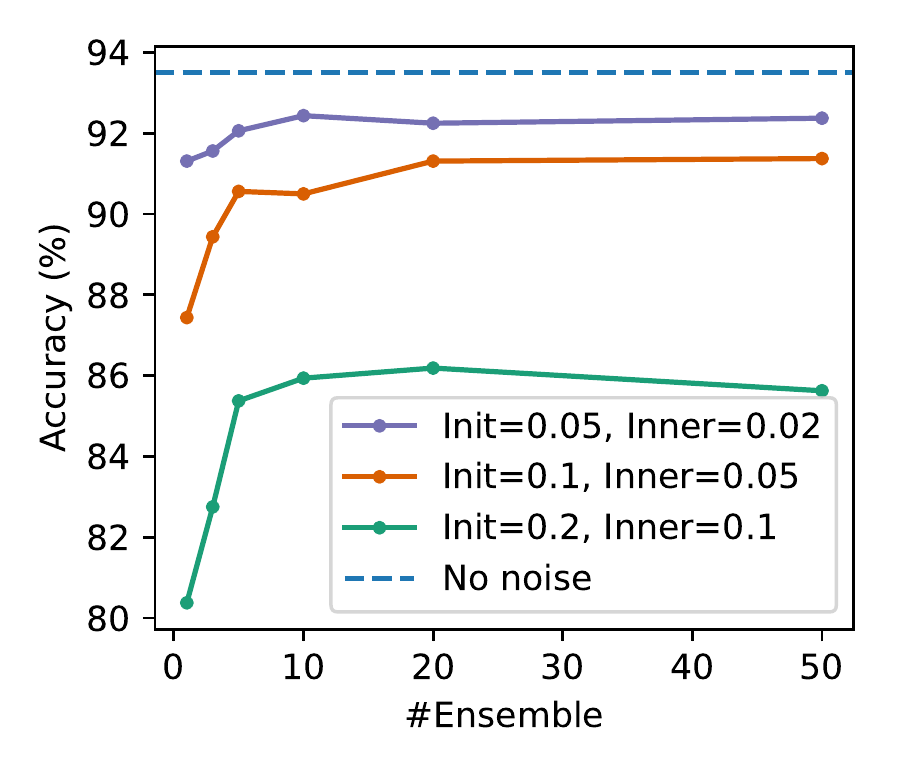}
    \caption{\textit{Left}: Comparing the accuracy under different levels of attack, here we choose VGG16+CIFAR10 combination. We can see that the ensemble model achieves better accuracy under weak attacks.  \textit{Right}:\label{fig:n-ensemble} Testing accuracy (without attack) of different $n$ (number of random models used for ensemble).}
    \label{fig:acc_ensemble}
\end{figure}
\begin{table}[tb]
\centering
\caption{Experiment setting for defense methods}
\label{tab:compare-set}
\begin{tabular}{@{}ll@{}}
\toprule
Methods                & Settings                                            \\ \midrule
No defense             & Baseline model                                           \\
RSE(for CIFAR10 + VGG16)                    & Initial noise: 0.2, inner noise: 0.1, 50-ensemble   \\
RSE(for CIFAR10 + ResNeXt) & Initial noise: 0.1, inner noise 0.1, 50-ensemble \\ 
RSE(for STL10 + Model A)                    & Initial noise: 0.2, inner noise: 0.1, 50-ensemble   \\
Defensive distill      & Temperature = 40                                    \\
Adversarial training (I) & FGSM adversarial examples, $\epsilon\sim\mathcal{U}(0.1, 0.3)$ \\
Adversarial training (II) & Following \cite{madry2017towards}, PGD adversary with $\epsilon_{\infty}=\frac{8.0}{256}$ \\
Robust Opt. + BReLU    & Following~\cite{zantedeschi2017efficient}                                      \\ \bottomrule
\end{tabular}
\end{table}
%\begin{figure}[tb]
%    \centering
%    \includegraphics[width=0.5\linewidth]{n_ensemble.pdf}
%    \caption{\label{fig:n-ensemble}Testing accuracy (without attack) of different $n$ (number of random models used for ensemble).}
%\end{figure}
\begin{table}[tb]
    \centering
      \caption{Prediction accuracy of defense methods under C\&W attack with different $c$. 
    We can clearly observe that  RSE is the most robust model. Our accuracy level remains at above 75\% when other methods are below 30\%.}
    \label{tab:my_label}
    \begin{tabular}{@{}cccccccc@{}}
    \toprule
           &$c=0.01$ & $c=0.03$ & $c=0.06$ & $c=0.1$ & $c=0.2$\\
           \midrule
       RSE(ours) &          \textbf{90.00\%}  & \textbf{86.06\%} & \textbf{79.44\%} & \textbf{67.19\%} & \textbf{34.75\%}  \\
    Adv retraining    & 27.00\% & 9.81\% & 4.13\% & 3.69\% & 1.44\% \\ 
    Robust Opt+BReLU  & 75.06\% & 47.93\% & 30.94\% & 20.69\% & 13.50\%\\
    Distill   & 49.88\%  & 17.69\% & 4.56\% &   3.13\% & 1.44\%\\
    No defense  & 30.38\%    & 8.93\%  & 5.06\%  & 3.56\%  & 2.19\% \\
    \bottomrule
    \end{tabular}
  
\end{table}

\begin{figure}[tb]
    \centering
    \includegraphics[width=0.9\linewidth]{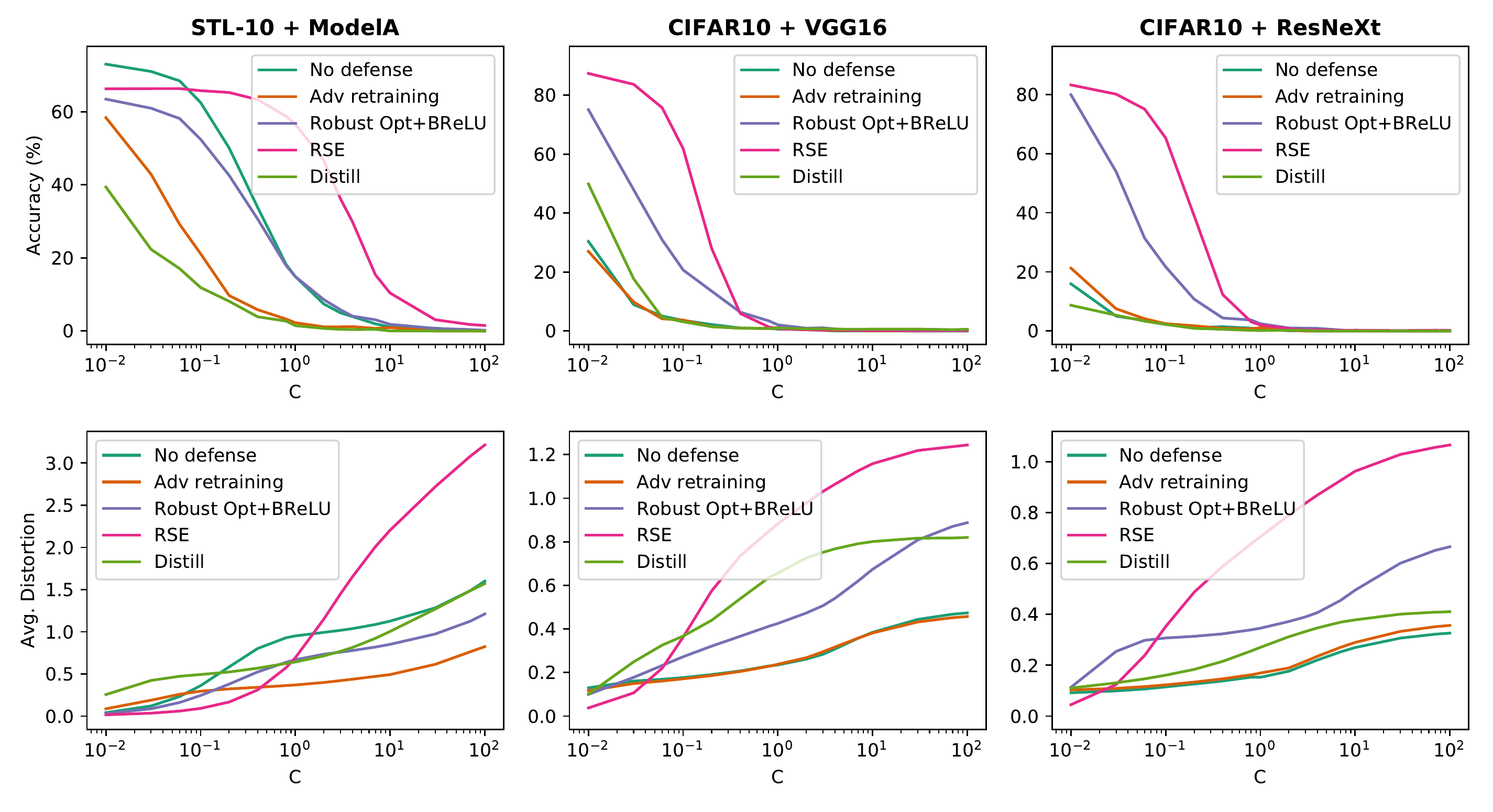}
    \caption{Comparing the accuracy of CIFAR10+\{VGG16, ResNeXt\} and STL10+Model A. We show both the change of accuracy and average distortion w.r.t. attacking strength parameter $c$ (the parameter in the C\&W attack). Our model (RSE) clearly outperforms all the existing methods under strong attacks in both accuracy and average distortion.}
    \label{fig:compare-defense}
\end{figure}

\begin{figure}[tb]
    \centering
    \includegraphics[width=0.8\linewidth,valign=m]{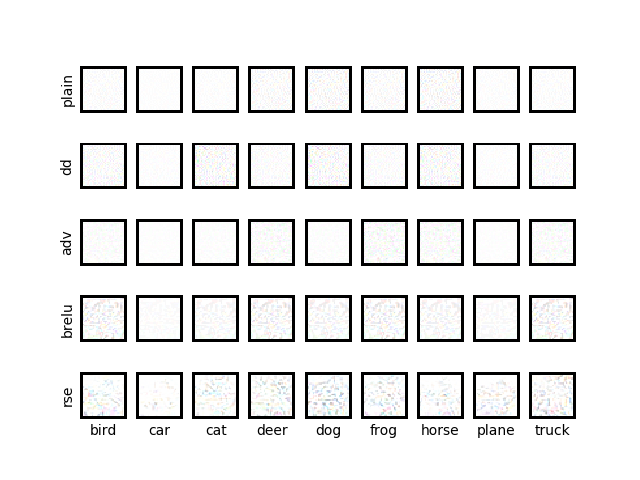}
    \includegraphics[width=0.1\linewidth,valign=m]{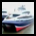}
    \caption{Targeted adversarial image distortion, each column indicates a defense algorithm and each row is the adversarial target (the original image is in ``ship'' class, shown in the right side). Here we choose $c=1$ for targetd C\&W attack. Visually, color spot means the distortion of images, thus a successful defending method should lead to more spots.}
    \label{fig:target-diff}
\end{figure}

\begin{table}[tb]
    \centering
    \begin{tabular}{|c|c|c|c|c|c|c|c|c|c|}
    \hline
         &  bird & car & cat & deer & dog & frog & horse & plane & truck\\\hline
    No defense&  1.94     & 0.31    & 0.74    & 4.72     &  7.99 & 3.66   & 9.22     & 0.75      & 1.32   \\\hline
    Defensive distill& 6.55      & 0.70    & \textbf{13.78}     & 2.54      & 13.90     & 2.56      & \textbf{11.36}       & 0.66      & 3.54    \\\hline
    Adv. retraining& 2.58      & 0.31 & 0.75    & 6.08     & 0.75     & 9.01     & 6.06      & 0.31       & 4.08   \\\hline
    Robust Opt. + BReLU& \textbf{17.11}      & 1.02     & 4.07     & 13.50     & 7.09    & 15.34     & 7.15      &  2.08     & 17.57   \\\hline
    RSE(ours)& 12.87      & \textbf{2.61}     & 12.47     & \textbf{21.47}      & \textbf{31.90}    & \textbf{19.09}      & 9.45       & \textbf{10.21}      & \textbf{22.15}   \\\hline
    \end{tabular}
    \caption{Image distortion required for targeted attacks. }
    \label{tab:distortion}
\end{table}

\subsection{The effect of noise level}
We first test the performance of RSE under different noise levels. We use Gaussian noise for all the noise layers in our network and the standard deviation $\sigma$ of Gaussian controls the noise level. 
Note that we call the noise layer before the first convolution layer the ``init-noise'', and all other noise layers the ``inner-noise''.
\par
In this experiment, we apply different noise levels in both training and testing phases to see how different variances change the robustness as well as generalization ability of networks. As an example, we choose 
\begin{equation}
    (\sigma_{\text{init}}, \sigma_{\text{inner}})=\{(0, 0), (0.05, 0.02), (0.1, 0.05), (0.2, 0.1)\}
\end{equation}
on VGG16+CIFAR10. The result is shown in Fig.~\ref{fig:noise-level-robust} (\textit{left}).

As we can see, both ``init-noise'' and ``inner-noise'' are beneficial to the robustness of neural network, but at the same time, one can see higher noise reduces the accuracy for weak attacks ($c\lesssim 0.01$). 
From Fig.~\ref{fig:noise-level-robust}, we observe that if the input image is normalized to $[0, 1]$, then choosing $\sigma_{\text{init}}=0.2$ and $\sigma_{\text{inner}}=0.1$ is good. Thus we fix this parameter for all the experiments.  

\subsection{Self-ensemble}
Next we show self-ensemble helps to improve the test accuracy of our noisy mode.  As an example, we choose VGG16+CIFAR10 combination and the standard deviation of initial noise layer is $\sigma=0.2$, other noise layers is $\sigma=0.1$. We compare 50-ensemble with 1-ensemble (i.e. single model), and the result can be found in Fig.~\ref{fig:acc_ensemble}.

We find the 50-ensemble method outperform the 1-ensemble method by ${\sim}8\%$ accuracy when $c<0.4$. This is because when the attack is weak enough, the majority choice of networks has lower variance and higher accuracy. On the other hand, we can see if $c>1.0$ or equivalently the average distortion greater than $0.93$, the ensemble model is worse. We conjecture that this is because when the attack is strong enough then the majority of random sub-models make wrong prediction, but when looking at any individual model, the random effect might be superior than group decision. In this situation, self-ensemble may have a negative effect on accuracy.
\par
Practically, if running time is the primary concern, it is not necessary to calculate many ensemble models. In fact, we find the accuracy saturates rapidly with respect to number of models, moreover, if we inject smaller noise then ensemble benefit would be weaker and the accuracy gets saturated earlier. Therefore, we find $10$-ensemble is good enough for testing accuracy, see Fig.~\ref{fig:n-ensemble}.

\subsection{Comparing defense methods}

Finally, we compare our RSE method with other existing defense algorithms.
Note that we test all of them using C\&W untargeted attack, which is the most difficult 
setting for defenders.

The comparison across different datasets and networks can be found in Tab.~\ref{tab:my_label} and Fig.~\ref{fig:compare-defense}. As we can see, previous defense methods have little effect on C\&W attacks. For example, Robust Opt+BReLU~\cite{zantedeschi2017efficient} is useful for CIFAR10+ResNeXt, but the accuracy is even worse than no defense model for STL10+Model A. In contrast, our RSE method acts as a good defence across all cases. Specifically, RSE method enforces the attacker to find much more distorted adversarial images in order to start a successful attack. As showed in Fig.~\ref{fig:compare-defense}, when we allow an average distortion of $0.21$ on CIFAR10+VGG16, C\&W attack is able to conduct untargeted attacks with success rate $>99\%$. On the contrary, by defending the networks via RSE, C\&W attack only yields a success rate of ${\sim}20\%$. Recently, another version of adversarial training is proposed \cite{madry2017towards}. Different from ``Adversarial training (I)'' shown in Tab.~\ref{tab:compare-set}, it trains the network with adversaries generated by multiple steps of gradient descent (therefore we call it ``Adversarial training (II)'' in Tab.~\ref{tab:compare-set}). Compared with our method, the major weakness is that it takes ${\sim}10$ times longer to train a robust network despite that the result is only slightly better than our RSE, see Fig.~\ref{fig:noise-level-robust} (\textit{right}).
\par
Apart from the accuracy under C\&W attack, we find the distortion of adversarial images also increases significantly, this can be seen in Fig.~\ref{fig:noise_ensemble}(2nd row), as $c$ is large enough (so that all defense algorithms no longer works) our RSE method achieves the largest distortion. 

Although all above experiments are concerning untargeted attack, it does not mean targeted attack is not covered, as we said, targeted attack is harder for attacking methods and easier to defense. As an example, we test all the defense algorithms on CIFAR-10 dataset under targeted attacks. We randomly pick an image from CIFAR10 and plot the perturbation $x_{\text{adv}}-x$ in Fig.~\ref{fig:target-diff} (the exact number is in Tab.~\ref{tab:distortion}), to make it easier to print out, we subtract RGB channels from 255 (so the majority of pixels are white and distortions can be noticed). One can easily find RSE method makes the adversarial images more distorted.
\par
Lastly, apart from CIFAR-10, we also design an experiment on a much larger data to support the effectiveness of our method even on large data. Due to space limit, the result is postponed to appendix.

\vspace{-5pt}\section{Conclusion}
\vspace{-5pt}
In this paper, we propose a new defense algorithm called Random Self-Ensemble (RSE) to improve the robustness of deep neural networks against adversarial attacks. We show that our algorithm is equivalent to ensemble a huge amount of noisy models together, and our proposed training process ensures that the ensemble model can generalize well. We further show that the algorithm is equivalent to adding a Lipchitz regularization and thus can improve the robustness of neural networks. Experimental results demonstrate that our method is very robust against strong white-box attacks. Moreover, our method is simple, easy to implement, and can be easily embedded into an existing network. 

\subsubsection*{Acknowledgement.}
The authors acknowledge the support of NSF via IIS-1719097 and the computing resources provided by Google cloud and Nvidia. 

\bibliographystyle{splncs04}
\bibliography{egbib}
\end{document}